\documentclass[10pt,twocolumn,letterpaper]{article}

\usepackage{cvpr}              
\definecolor{cvprblue}{rgb}{0.21,0.49,0.74}
\usepackage[pagebackref,breaklinks,colorlinks,allcolors=cvprblue]{hyperref}

\def\confName{CVPR}
\def\confYear{2027}

\usepackage{amsmath,amsfonts,amssymb}
\usepackage{algorithmic}
\usepackage{algorithm}
\usepackage{array}
\usepackage{caption}
\usepackage{subcaption}
\usepackage{textcomp}
\usepackage{stfloats}
\usepackage{url}
\usepackage{verbatim}
\usepackage{graphicx}
\usepackage{comment}
\usepackage{color}
\usepackage{bbm}
\usepackage{multirow}
\usepackage{booktabs}
\usepackage[table]{xcolor}

\title{ReMoA: Recursive Modular Adaptation via Sparse Latent Refinement in \\ Multimodal Diffusion Models}

\author{%
  \normalsize Yuwei Sun\textsuperscript{1}, Yuxuan Yao\textsuperscript{2}, Hui Li\textsuperscript{2}, Weihao Yuan\textsuperscript{3}, Siyu Zhu\textsuperscript{1,2} \\
  \normalsize\textsuperscript{1}Shanghai Academy of AI for Science, 
  \textsuperscript{2}Fudan University,
  \textsuperscript{3}Nanjing University
}

\begin{document}
\maketitle

\begin{abstract}
    Diffusion models have achieved success in high-fidelity data synthesis, yet their capacity for more complex, structured reasoning such as text following remains constrained. While advances in language models have leveraged strategies such as latent reasoning and recursion to enhance text understanding capabilities, extending these to multimodal text-to-image generation tasks is challenging due to the continuous and non-discrete nature of visual tokens. To tackle this problem, we draw inspiration from modular human cognition and propose a Recursive Modular Adaptation (ReMoA) framework integrated into joint attention layers of conventional diffusion models, iteratively refining visual tokens over multiple recursion steps while efficiently sharing parameters via sparse selection of neural modules. Comprehensive evaluations on both class-conditioned ImageNet generation and text-to-image generation using the GenEval and DPG benchmarks demonstrate that recursive visual latent refinement contributes to richer object details and better-aligned outputs. Additionally, visual latent trajectory and routing analyses reveal that neural modules learn to specialize across diffusion timesteps, with each module's trajectories diverging at later, low-noise steps to support context-adaptive refinement. ReMoA increases recall for DiT-XL from 0.57 to 0.62 on ImageNet, and improves the overall GenEval score for SD3-medium from 68.33\% to 71.18\%, without introducing additional parameters through recursion.
\end{abstract}


\section{Introduction}

Diffusion models have achieved remarkable success in generating high-fidelity, diverse outputs. However, while these models excel at data synthesis, their capacity for structured, multi-step reasoning for enhanced text-following capability remains underexplored. This gap mirrors a long-standing challenge in the language domain, where recent advances in large language models (LLMs) such as Chain-of-Thought (CoT) and Test-time Latent Reasoning solve complex problems by either generating explicit intermediate reasoning steps or recursive internal state updates \cite{wei2022chain,wang2022rationale,snell2024scaling}. Nevertheless, progress in iterative latent refinement has been largely confined to language, with limited exploration in multimodal scenarios such as text-to-image generation tasks.

One obstacle is that, unlike language tokens, visual information is less discrete and cannot be efficiently reasoned over or is simply computationally expensive and data-hungry within its continuous space. To tackle this problem, we draw inspiration from the modular and adaptive nature of human multimodal cognition. We build on recent work on sparse transformers \cite{madan2021fast, zhu2025llada, zhao2024dynamic, sun2025associative} and propose a Recursive Modular Adaptation (ReMoA) framework tailored to diffusion models for better capturing complex patterns within the diffusion latent space through recursive latent refinement. We start with class-conditioned Diffusion Transformers (DiTs) and extend the architecture to multimodal Stable Diffusion 3 (SD3) model. We aim to demonstrate that recursive latent computation could also benefit vision tasks, efficiently improving generation quality and text–visual alignment.

Our approach centers on a recursive component integrated into attention layers. This component iteratively refines visual tokens over multiple latent steps using a set of shared neural modules. These modules are sparsely activated at each latent step and dynamically selected by a gating network conditioned on the current vision tokens and on auxiliary conditioning signals, namely, the class label, the diffusion timestep, and the text prompt embeddings. We further incorporate two mechanisms to ensure learnable expert routing and efficient recursive training: (1) a Gumbel-Softmax strategy \cite{lei2024compete} that enables winner-takes-all gradient allocation for training the sparse selection component, encouraging the emergence of distinct neural modules, and (2) a tailored mixture-of-adapters mechanism with recursively updated modules that efficiently accumulates the outputs of successive latent steps.

We evaluate the performance of ReMoA on both class-conditioned generation on ImageNet \cite{deng2009imagenet}, measured by FID, sFID, IS, precision, and recall, and the more complex text-to-image generation on the GenEval \cite{ghosh2023geneval} and DPG \cite{hu2024ella} benchmarks. Comprehensive empirical evaluations, including ablation studies and latent trajectory visualizations, demonstrate the benefits of the proposed method in improving generation quality and text–visual alignment.

Overall, our main contributions are threefold:

1) We propose a recursive modular adaptation framework specifically tailored to vision diffusion models, including DiTs and SD3. The framework enhances conditional alignment in image generation via iterative modular processing of vision tokens with a condition-guided routing strategy (Section \ref{sec:module}, \ref{sec:recursion}).

2) Latent trajectory and routing analyses show that visual latents evolve coherently early in diffusion but diverge at later, low-noise timesteps, revealing that neural modules progressively specialize and facilitate diverse activation patterns across recursive steps (Section \ref{sec:traj}).

3) Extensive experiments on ImageNet and text-to-image generation tasks on the GenEval and DPG benchmarks reveal efficient model adaptation and progressive enhancement in image generation quality and text-visual alignment (Section \ref{sec:t2i}).

\section{Related work}

This section provides a comprehensive summary of the most relevant recent work on diffusion transformers, sparse modular computation, and latent reasoning methods.

\paragraph{Diffusion Transformers}

Diffusion models \cite{ho2020denoising,song2020denoising} have become the foundation of modern image generation, enabling high-fidelity and diverse synthesis \cite{rombach2022high}. A key architectural shift was pioneered by Diffusion Transformers (DiTs) \cite{peebles2023DIT}, which demonstrated the scalability and effectiveness of Transformer backbones for diffusion and inspired a series of follow-up works on improved conditioning mechanisms \cite{chen2023pixartalpha,chen2024pixartsigma,zhou2025goldennoisediffusionmodels}. This line of work culminates in joint multimodal architectures such as Stable Diffusion 3 (SD3) \cite{esser2024sd3}, an instantiation of Multimodal Diffusion Transformers (MMDiTs). Despite these advances, however, existing architectures process conditioning information in a largely monolithic manner, leaving room for more structured, modular processing during denoising.

\paragraph{Modularity and sparse transformers}

Unlike the monolithic design of conventional MMDiTs, cross-modal cognition in humans typically relies on sparse activations through a communication bottleneck \cite{goyal2021coordination,blum2021theoretical}. Inducing sparsity in Transformer-based architectures via modulation has led to the emergence of distinct functions in neural modules, enhancing both performance and parameter efficiency \cite{madan2021fast,lei2024compete,sun2025associative}. For example, LLaDA-MoE \cite{zhu2025llada} integrates a mixture-of-experts architecture into the training objective of masked diffusion models. Dynamic Diffusion Transformer \cite{zhao2024dynamic,zhao2025dydit++} further showed emergent resource allocation across neural modules, with some modules employed more frequently in earlier layers and less frequently in later layers. Inspired by these findings, we explore sparsely activated modules to iteratively refine diffusion latents, improving conditioning alignment over multiple steps.

\paragraph{Latent reasoning}

Computational depth has proven valuable for reasoning tasks that require multi-step compositional thinking. Latent reasoning, iterative refinement within a latent space, has been investigated primarily in language models. For instance, Pondering Language Models \cite{banino2021pondernet} perform iterative ponder cycles within each token prediction, where weighted hidden states are fed back through a residual path for refinement. Tiny Networks \cite{jolicoeur2025less} and the Hierarchical Reasoning Model (HRM) \cite{wang2025hierarchical} employ a hierarchical reasoning framework to recursively refine latent states for text-based puzzle tasks. Test-time Latent Reasoning \cite{geiping2025scaling}, using a random number of latent steps, demonstrates advantages in enhancing performance.

Latent refinement has recently been extended to multimodal models beyond language. For instance, the Chain of Continuous Vision-Language Thought (CoCoVa) \cite{ma2025cocova} refines a chain of latent states through gated cross-modal fusion, while the Multimodal Chain of Continuous Thought (MCOUT) \cite{pham2025multimodal} enables reasoning in a joint latent space, with reasoning states iteratively refined and aligned. Despite these advances, leveraging latent reasoning to enhance vision-text alignment in conditioned image generation tasks with diffusion models remains largely unexplored. Our work addresses this gap by proposing a recursive modular adaptation mechanism built into the joint attention layers of diffusion models.

\section{Method}

\begin{figure*}[t]
    \centering
    \includegraphics[width=\linewidth]{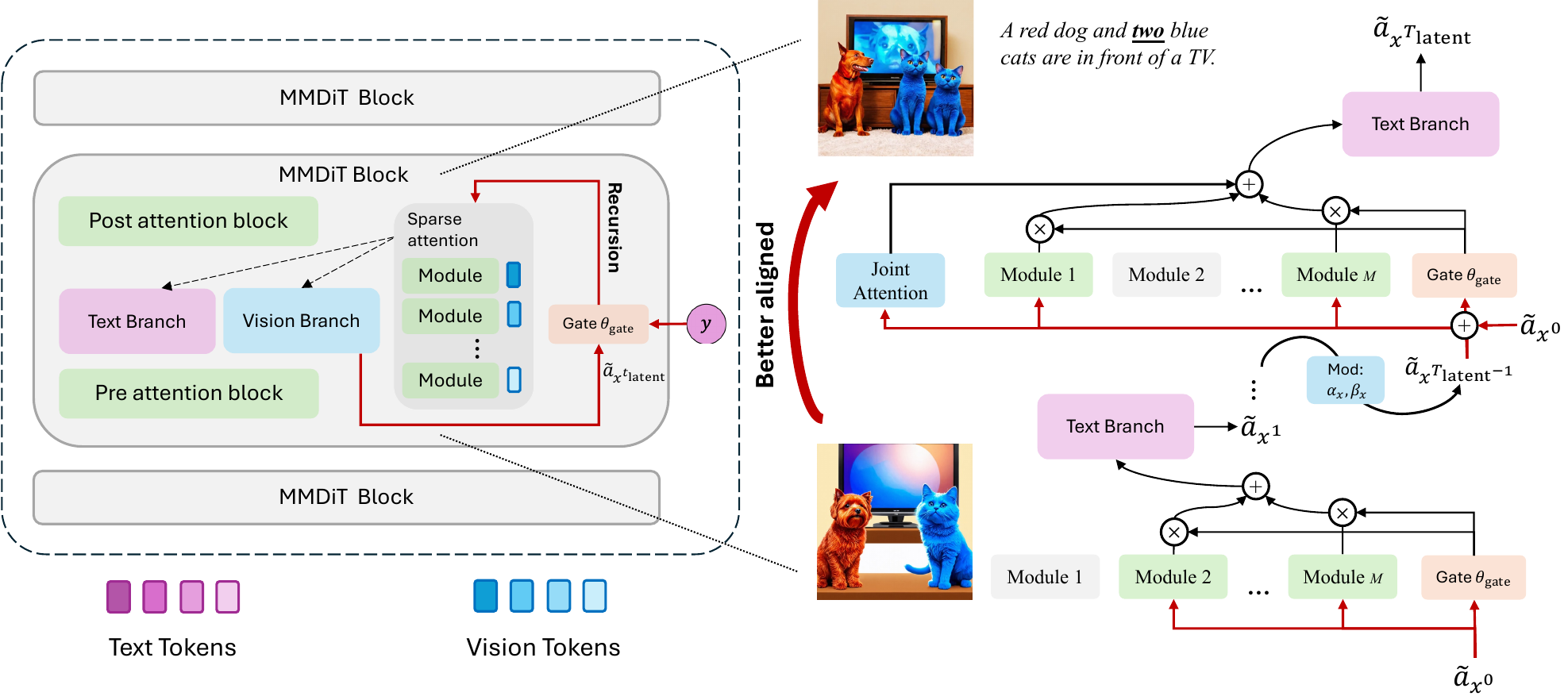}
    \caption{Architecture of the proposed Recursive Modular Adaptation (ReMoA) mechanism integrated with SD3. A recursive modular component is inserted into the joint attention of the vision and text branches: routing probabilities are computed per visual token across samples, and tokens are reassembled into their original spatial order after passing through the selected modules. At each latent step, the module outputs interact with the text branch to progressively refine the visual latents under text guidance, yielding generation better aligned with the text prompt.}
    \label{fig:scheme}
\end{figure*}

This section describes the key technical underpinnings of the Recursive Modular Adaptation (ReMoA) mechanism, involving a mixture-of-adapters tailored to diffusion transformers and a recursive sparse joint attention mechanism (Figure \ref{fig:scheme}). Introducing such an architecture into Multimodal Diffusion Transformer (MMDiT) blocks, specifically those responsible for structural layout, can help resolve prompt ambiguities and enhance semantic alignment in generated images. Notably, we induce recursive modular processing via a set of neural modules sparsely activated over several latent steps, conditioned on the current vision tokens, the diffusion timestep, and the text prompt embeddings. For concreteness, we focus on the MMDiT model of SD3 in this section, where the model progressively refines the alignment between textual and visual features via joint attention. In addition, we validate ReMoA on Diffusion Transformers (DiTs) and defer the detailed configurations to the experimental section.

\subsection{Multimodal Diffusion Transformers}

Multimodal Diffusion Transformers (MMDiTs) process the vision and text tokens through two individual flows and employ joint attentions to unify the connections among the different modalities. Within each block, self-attention is first utilized intra-modally to refine the representations within each modality independently. A joint attention mechanism then models cross-modal dependencies, typically by projecting the concatenated sequence of vision and text tokens through a unified attention operation. This allows every token, regardless of modality, to attend to all others, effectively enabling direct interactions between visual and textual elements. 

In particular, let $x$ be the vision branch input, $c$ be the text branch input, and $y$ be embeddings derived from the diffusion timestep and sinusoidal encoding of text prompts. Each block consists of adaptive normalization, joint attention, and MLP layers. 
We target the joint attention component in MMDiT blocks, where the normalized tokens are concatenated as $h = [\tilde{x};\tilde{c}]$ and used as the input to a multi-head attention mechanism Attn($\cdot$) as follows:
\begin{align}
Q = h W_Q, \quad K = h W_K, \quad V = h W_V,\\
\mathrm{Attn}(h) = \mathrm{Softmax}\Big(\frac{Q K^\top}{\sqrt{D}}\Big) V,
\end{align}
where \( W \in \mathbb{R}^{D \times D} \) for \( W \in \{W_Q, W_K, W_V\} \) and \( D \) is the dimension of the weight matrices. After the attention, the outputs are split back into the image and text streams, i.e., \([a_x;a_c] = \mathrm{Attn}(h)\). Then, the attention outputs are added to the original tokens from each modality as residuals, with learnable gating parameters \(\gamma_x\) and \(\gamma_c\) as follows:
\begin{equation}
\begin{aligned}
x' &= x + \gamma_x \cdot A_x,\\
c' &= c + \gamma_c \cdot A_c.
\end{aligned}
\end{equation}

Finally, the residual outputs are passed through LayerNorm (LN) and an MLP with residual gating. Consequently, a complete single MMDiT block can be summarized as follows:
\begin{equation}
\begin{aligned}
h &= [\tilde{x}, \tilde{c}],\\
h &\leftarrow h + \gamma \cdot \mathrm{Attn}(\alpha \cdot \mathrm{LN}(h) + \beta), \\
h &\leftarrow h + \zeta \cdot \mathrm{MLP}(\delta \cdot \mathrm{LN}(h) + \epsilon),
\end{aligned}
\end{equation}
where $\alpha, \beta, \gamma, \delta, \epsilon, \zeta$ are modulation parameters separate for each modality $x$ and $c$.

\subsection{Neural modules with mixture-of-adapters}
\label{sec:module}

We introduce several adapters into the joint attention architecture that dynamically process visual tokens and remap them back to their original spatial order. These visual tokens are recursively updated across latent steps, with each step passing them through distinct neural modules selected by a gating network to progressively refine the representation. This design creates a competitive bottleneck, forcing the neural modules to develop specialized processing functions as they compete for selection by the gating network, integrating the vision tokens assigned to them under the guidance of the text prompt embeddings, thereby enhancing cross-modal alignment at each step.

To induce a recursive component in diffusion models, a naive choice is a monolithic model that takes its own output back to the input via layer repetition. Nevertheless, this design could be inefficient for recursive training, since the shared component goes through multiple iterations and a monolithic model could significantly increase training cost. Thus, we opt for a more efficient mixture-of-adapter architecture with a set of lightweight neural modules sparsely activated for each iteration. In particular, the recursive component $\theta_{\text{recursive}}$ consists of a total of \( M \) neural modules $\{\theta_{\text{adapter}}^m\}_{m=1}^M$ and a gating network \( \theta_\text{gate} \). To determine which module to activate, the gating network produces selecting probabilities for $M$ modules. To increase the capacity of neural modules for learning distinct features, we compute the probability for each visual token $v$ independently across samples. Then, after each latent step, we reassemble the tokens into their original spatial order within each image. This allows various modules to access tokens across samples, improving learning efficiency, while the reassembly process preserves the original spatial arrangement after each latent recursion step. Furthermore, the routing probability is computed based on (1) the current latent step's vision token \( x^{t_\text{latent}} \), (2) the diffusion timestep, and (3) the conditioning text prompt embeddings. Note that, in MMDiTs, the text embeddings and the diffusion timestep is usually combined and represented as $y$. Finally, for each step, only one module is selected and activated, given by:
\begin{equation}
\theta_{\text{adapter}}^{m^*} : \arg\max_m\, P_m^{t_\text{latent}}(\theta_\text{gate}(x^{t_\text{latent}}, y)),
\end{equation}
where \( P_m^{t_\text{latent}} \) denotes the selection probability for module \( m \). In practice, we employ the Gumbel-Softmax to enable gradients back-propagated through the recursive component, for the hard selection of a single module. Thus, we compute logits $\hat{z}^{t_\text{latent}}_m $ for the module selection as follows:
\begin{align}
z_m^{t_\text{latent}} &= P_m^{t_\text{latent}} + \epsilon^{t_\text{latent}},\\ 
\hat{z}^{t_\text{latent}}_m &= \text{softmax}\left( z^{t_\text{latent}}_m / \tau \right),\\
\theta_{\text{adapter}}^{t_\text{latent}} &: \arg\max_m\, \hat{z}^{t_\text{latent}}_m,
\end{align}
where \( \epsilon^{t_\text{latent}} \sim \text{Gumbel}(0,1) \) adds noise for exploration during training and the temperature \( \tau \) controls the sparsity of selection.

\subsection{Recursive sparse joint attention}
\label{sec:recursion}


In an MMDiT block, we target the joint attention component where the two modalities exchange information, as the alignment between vision and text modalities is built at this stage. We employ an adaptation scheme that recursively implements Low-Rank Adaptation (LoRA) \cite{hu2021lora} across \( R \) latent steps to construct lightweight neural modules for efficient adaptation. In particular, for each joint attention component, trainable low-rank updates are applied to the query, key, and value projection matrices \( W^Q_{\text{vision}}, W^K_{\text{vision}}, W^V_{\text{vision}} \in \mathbb{R}^{D \times D}  \) of the vision branch. For a weight matrix \( W \in \{W^Q_{\text{vision}}, W^K_{\text{vision}}, W^V_{\text{vision}}\}\), the update \( \Delta W \) is computed as \( \Delta W = BA \), where \( A \in \mathbb{R}^{r \times D} \), \( B \in \mathbb{R}^{D \times r} \), and the rank \( r \ll D \). Thus, for an input vision token \( x^{t_{\text{latent}}} \), the output becomes $\hat{W} x^{t_{\text{latent}}} + \Delta W x^{t_{\text{latent}}} = \hat{W} x^{t_{\text{latent}}} + BA x^{t_{\text{latent}}}$, where \( \hat{\cdot} \) denotes frozen parameters.

In initial experiments, we found that processing the input through the frozen base model repeatedly at each step corrupted the generation performance, as repeated exposure to the base model's fixed representations introduced compounding artifacts (Figure \ref{fig:lora}). To this end, we leverage the low-rank adapter outputs for recursive computation throughout the intermediate latent steps, and the frozen base model output is integrated at the final step. This design prevents the results of recursive computation from drifting significantly away from the original distribution space toward over-smoothed predictions, while enabling progressive visual refinement through the recursive adaptation.

At each latent step \( t_\text{latent} \in \{1, \dots, T_\text{latent}\} \), a specific set of LoRA parameters \( \theta_{\text{adapter}}^{t_\text{latent}}: \{B^{t_\text{latent}}, A^{t_\text{latent}}\} \) is selected based on the predicted routing probability \( P_m^{t_\text{latent}} \). We then compute the updated vision tokens, concatenate them with the text tokens \(\tilde{c}\) for joint attention, and split the computed result back into the vision and text branches. Notably, we leverage the normalized visual tokens $\tilde{x}$ as the initial input of the visual branch, i.e., $\tilde{a}_{x^{0}} = \tilde{x}$, and recursively compute the adaptation results at various intermediate latent steps $t_\text{latent}= 1, \dots, T_\text{latent}-1$ as follows:
\begin{align}
\tilde{a}_{x^{t_\text{latent}-1}}^\text{AdaLN} &= \alpha_x\cdot \text{LN}(\tilde{a}_{x^{t_\text{latent}-1}})+\beta_x, \\
 [a_{x^{t_\text{latent}}};a_{c^{t_\text{latent}}}] &= \text{Attn}(B^{t_\text{latent}} A^{t_\text{latent}}\,\tilde{a}_{x^{t_\text{latent}-1}}^\text{AdaLN};\tilde{c}),\\
\tilde{a}_{x^{t_\text{latent}}} &= a_{x^{t_\text{latent}}} + \tilde{x}.
\end{align}
Finally, the recursive adaptation after \( T_\text{latent} \) steps incorporates the output of the frozen base model to produce the output of the sparse joint attention component, i.e., $[a_{x^{T_\text{latent}}};a_{c^{T_\text{latent}}}] = \text{Attn}(B^{t_\text{latent}} A^{t_\text{latent}}\, (\alpha_x\cdot \text{LN}(\tilde{a}_{x^{T_\text{latent}-1}})+\beta_x) +\hat{W}\tilde{x};\tilde{c})$, where $a_{x^{T_\text{latent}}}$ and $a_{c^{T_\text{latent}}}$ are the outputs of the recursive modular component for the vision and text branches, respectively.

\begin{figure*}
    \centering
    \includegraphics[width=\linewidth]{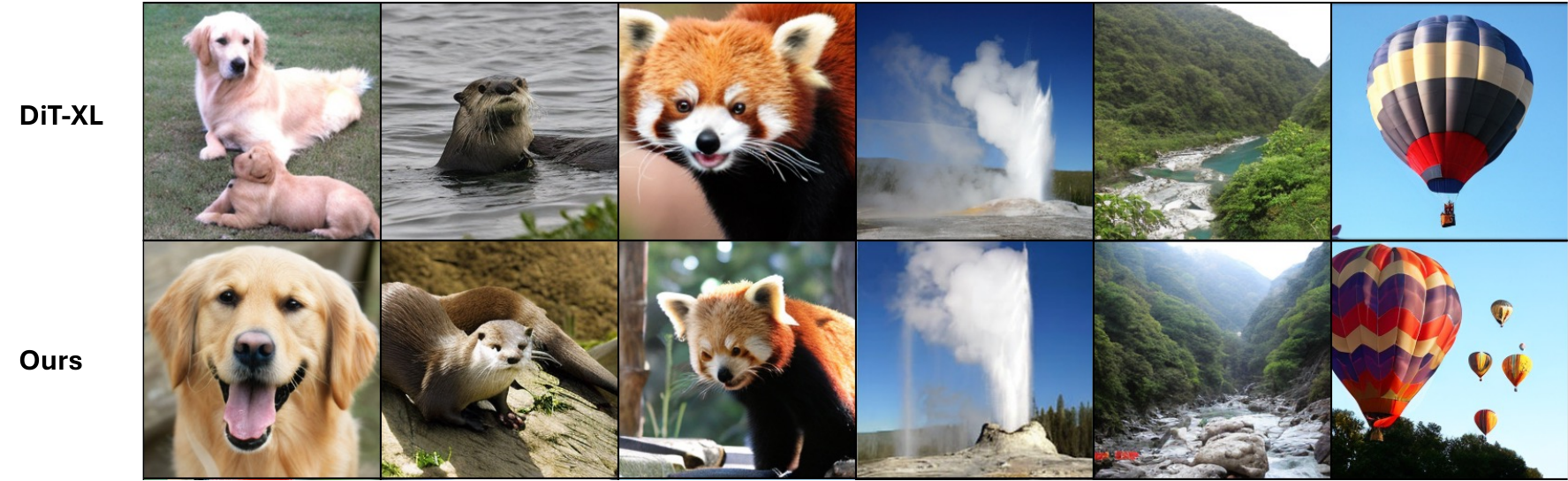}
    \caption{Image generation results of DiT-XL and the proposed method based on the recursive modular adaptation (ReMoA) mechanism. Using the recursive method provides richer object textures and finer background details in the generated images.}
    \label{fig:dit}
\end{figure*}

\begin{table}[t]
\small
\setlength{\tabcolsep}{4pt}
\centering
\renewcommand{\arraystretch}{1.2}
\caption{Performance comparison for class-conditioned image generation on the ImageNet dataset at \(256 \times 256\) resolution. ReMoA outperforms DiT-XL/2, particularly in recall (0.62 vs. 0.57), consistent with the recursive architecture's ability to produce more diverse and higher-quality generations. Best per column in \textbf{bold} and second-best \underline{underlined}.}
\begin{tabular}{lccccc}
\rowcolor{CornflowerBlue!10}
\toprule   
Model &  FID$\downarrow$ & sFID$\downarrow$ & IS$\uparrow$ & Precision$\uparrow$ & Recall$\uparrow$ \\
\midrule
ReMoA (Ours)   & \textbf{2.27}  & \textbf{4.69} & \textbf{275.64} & \underline{0.82} & \textbf{0.62} \\  
DiT-XL/2 \cite{peebles2023DIT} &  2.34 & \underline{4.73}  & \underline{275.56} & \textbf{0.83} & \underline{0.57} \\ 
\hline
\multicolumn{6}{l}{Other methods} \\
\hline
ADM \cite{dhariwal2021diffusion} & 4.59 & 5.25 & 186.87 & \underline{0.82} & 0.52 \\
LDM-4 \cite{rombach2022high} & 3.95 & --   & 178.22 & 0.81 & 0.55 \\
U-ViT-L/2 \cite{bao2023all}  & 3.40 & 6.63 & 219.94 & \textbf{0.83} & 0.52 \\
U-ViT-H/2 \cite{bao2023all}   & \underline{2.29} & 5.68 & 263.88 & \underline{0.82} & \underline{0.57} \\
\bottomrule
\end{tabular}
\label{tab:classcondition}
\end{table}

\section{Experiments}

In this section, we present the experimental settings and extensive empirical results. We first evaluate the proposed method on class-conditional ImageNet generation using Diffusion Transformers (DiTs), including a detailed recursive latent trajectory analysis. We then extend our evaluation to more complex text-following tasks based on Stable Diffusion 3 (SD3), assessing performance on the GenEval and DPG benchmarks with ablation studies and a discussion on the refinement method.

\subsection{Settings}

\paragraph{Datasets}

We consider both class-conditioned and text-to-image generation. For class-conditioned generation, we employ the ImageNet \cite{deng2009imagenet} dataset at a resolution of 256 × 256. For text-to-image generation, we evaluate on the GenEval benchmark \cite{ghosh2023geneval}, which measures various attributes including single object, two objects, counting, colors, color attribution, and position, and additionally report generation results on the DPG benchmark \cite{hu2024ella}. In both benchmarks, we fine-tune models with 1000 random samples from the MSCOCO dataset \cite{lin2015microsoftcococommonobjects}.

\paragraph{Model variants and hyperparameters}

We employ SD3-medium \cite{esser2024sd3} with a total of 24 layers, applying the recursive component in the joint attention layer of vision and text modalities. For class-conditioned generation, we employ a DiT-XL/2 model \cite{peebles2023DIT} with a total of 24 layers, applying the recursive component to the self-attention layer using the class labels and diffusion timestep as input to the gating network. For the DiT-XL/2 variant, we select the middle layer (layer 12) as the recursive layer, whereas for SD3-medium we apply the recursive component to either the middle layer (layer 12) or multiple layers (e.g., layers 4, 8, and 12). We train with the AdamW optimizer using a learning rate of $5 \times 10^{-4}$, betas of $(0.9, 0.999)$, and zero weight decay, with a batch size of 128. The LoRA rank is set to 128, and we set the number of experts and the latent recursion steps to 5. Additionally, we set the temperature parameter $\tau$ of the Gumbel Softmax to 5.0, with a top-k temperature of 0.1. DiT-XL/2 models are trained for 20 epochs at an input image size of 256, while SD3-medium is trained for 50 epochs at 1024 resolution. Unless otherwise noted, we employ the author-recommended settings and hyperparameters for the re-implementation of all baseline models.


\begin{figure*}[t]
\centering
\includegraphics[width=\linewidth]{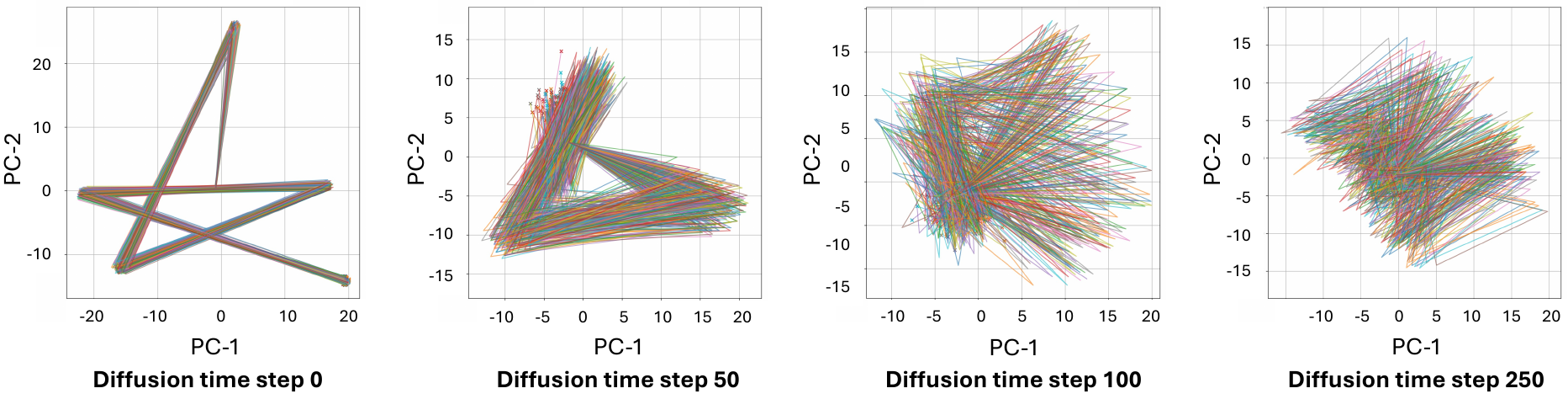}
\caption{Latent trajectories of vision tokens during five-step recursion (12th layer). Visualized via PCA compressing all tokens from a single image. Trajectory changes across diffusion timesteps: at early steps (high noise), trajectories are largely coherent, indicating uniform processing; at later steps (low noise), they exhibit divergent, patch-specific pathways, reflecting the neural modules' context-adaptive specialization.}
\label{fig:trajectories}
\end{figure*}

\begin{figure*}[t]
\centering
\includegraphics[width=0.85\linewidth]{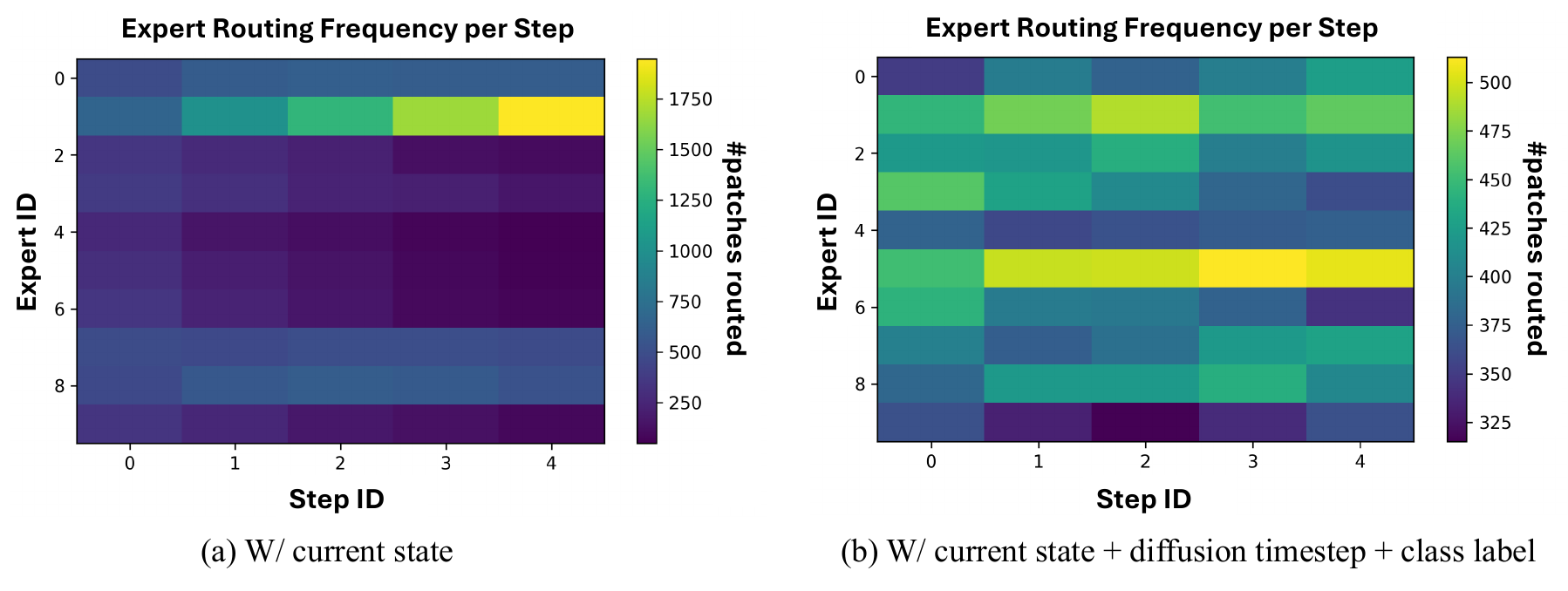}
\caption{Expert routing frequency per step. Adaptive neural module activation conditioned on the current vision tokens, the diffusion timestep, and the class label. Ablating the conditioning to only the current vision tokens results in modules that struggle to diversify across latent steps. Therefore, the optimal choice incorporates both the current states and the additional conditional information.}
\label{fig:routing_stats}
\end{figure*}

\subsection{Class-conditioned image generation}
\label{sec:class}

For class-conditioned generation, following \cite{teng2024dim}, we sample 50K images to compute the Fréchet Inception Distance (FID) \cite{heusel2017gans}, Inception Score (IS) \cite{salimans2016improved}, sFID \cite{nash2021generating}, precision, and recall. We compare against SD3-medium and relevant conventional generation models in Table \ref{tab:classcondition}. The proposed recursive framework outperforms DiT-XL/2, particularly in recall (0.62 vs. 0.57), consistent with the recursive architecture's ability to produce more diverse and higher-quality generations. In addition, generated image samples are shown in Figure \ref{fig:dit}, where the recursive method yields richer object textures and finer background details.

\paragraph{Expert allocation and latent trajectory analysis}
\label{sec:traj}

To understand the evolution of visual latents during the various recursion steps, we performed a more detailed analysis of the routing behavior during inference. As shown in Figure \ref{fig:trajectories}, the latent trajectories appear unified, with similar behavior early in the diffusion process but diverge at later diffusion timesteps, indicating increased neural module specialization. This reveals how architectural specialization emerges dynamically throughout the generative process. Furthermore, the dynamic routing results illustrated in Figure \ref{fig:routing_stats} demonstrate how neural module activation shifts across recursive steps, reflecting a progressive refinement of visual latents, while using the diffusion timestep, class label, and current vision tokens enables effective and diverse module activations.

\begin{figure*}[t]
    \centering
    \includegraphics[width=\linewidth]{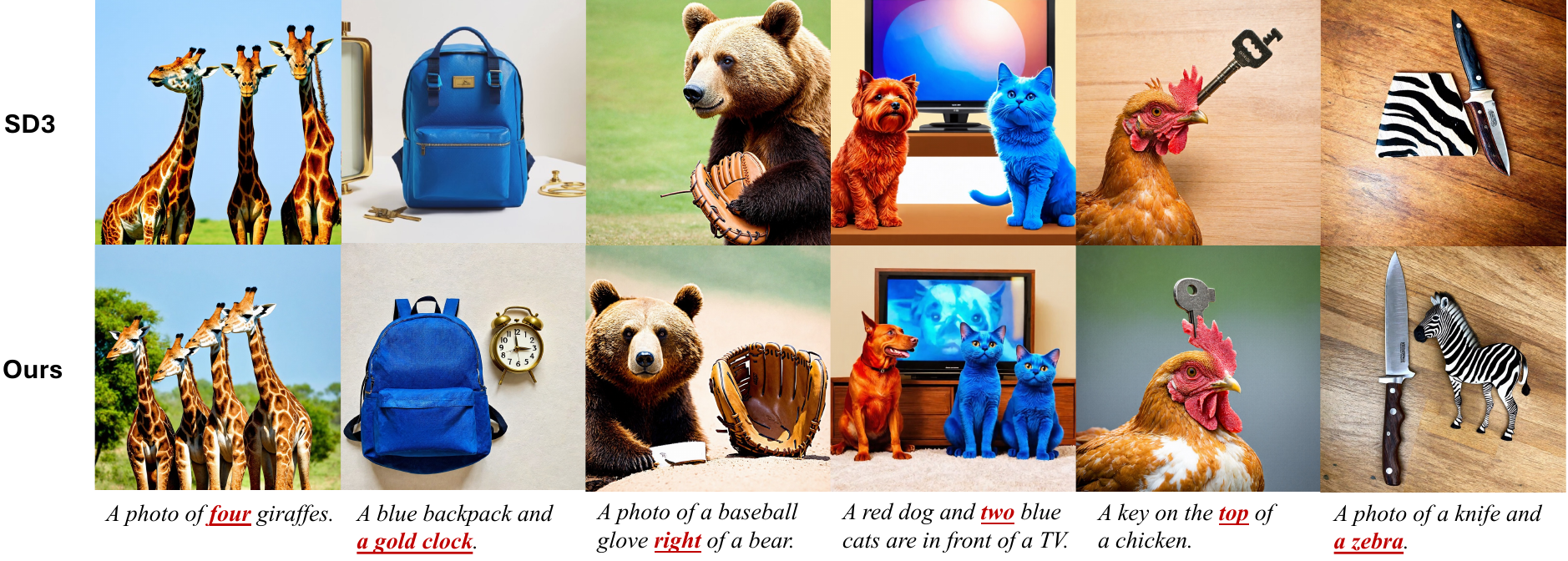}
    \caption{Qualitative comparison of image generations from our recursive approach versus the SD3-medium baseline at 1024 × 1024 resolution. Our method achieves higher text alignment and fidelity than the SD3 baseline, which maintains high sample fidelity but often fails to faithfully reflect fine-grained prompt details described in the text.}
    \label{fig:sd3-compare}
\end{figure*}

\begin{figure}[t]
    \centering
    \vspace{10pt}
    \includegraphics[width=\linewidth]{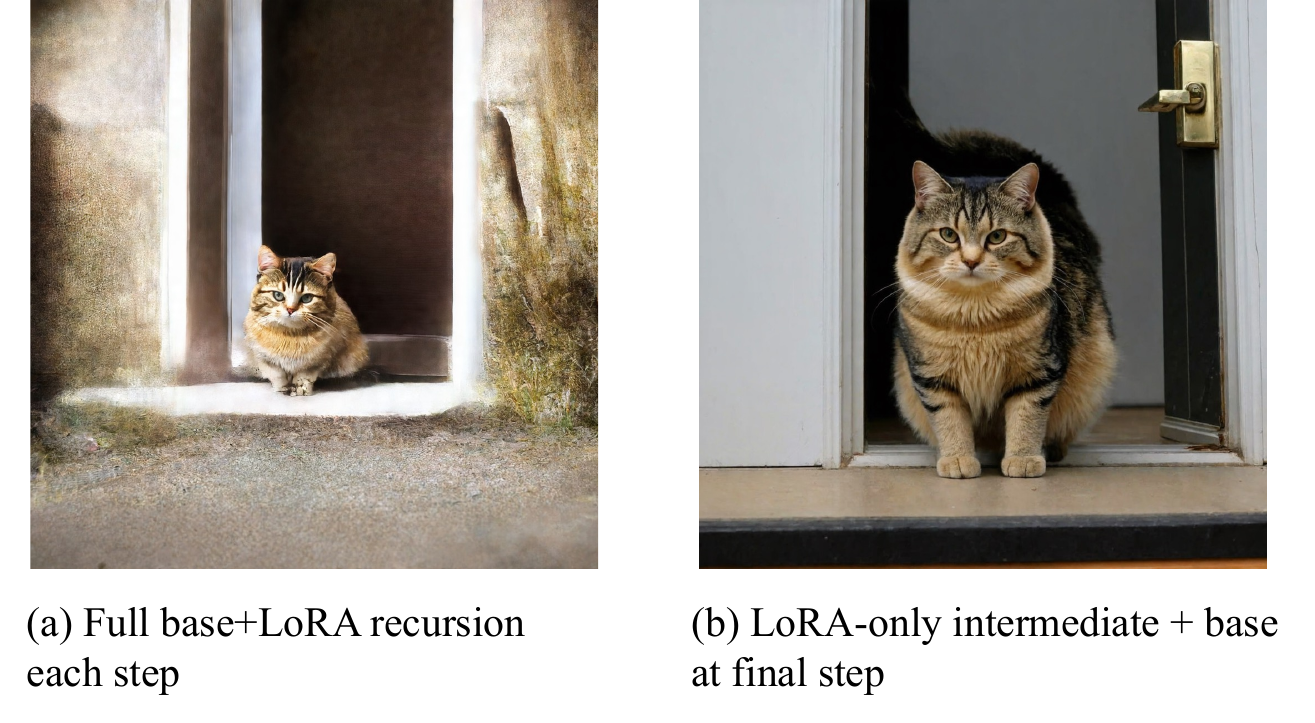}
    \caption{Injection point of the frozen base model's output during the recursive refinement, with the prompt fixed as “A cat is waiting in front of a door.” (a) Every-step injection: the frozen base projection is applied at every latent step, so that its fixed output is repeatedly fed back through the LoRA branch, causing background details to blur together and noisy artifacts to accumulate. (b) Final-step injection (ours): intermediate steps employ only the LoRA adapters, and the frozen base projection is integrated once at the final step, thereby preserving the pretrained distribution while still allowing progressive refinement.}
    \label{fig:lora}
\end{figure}

\begin{figure}[t]
    \centering
    \includegraphics[width=\linewidth]{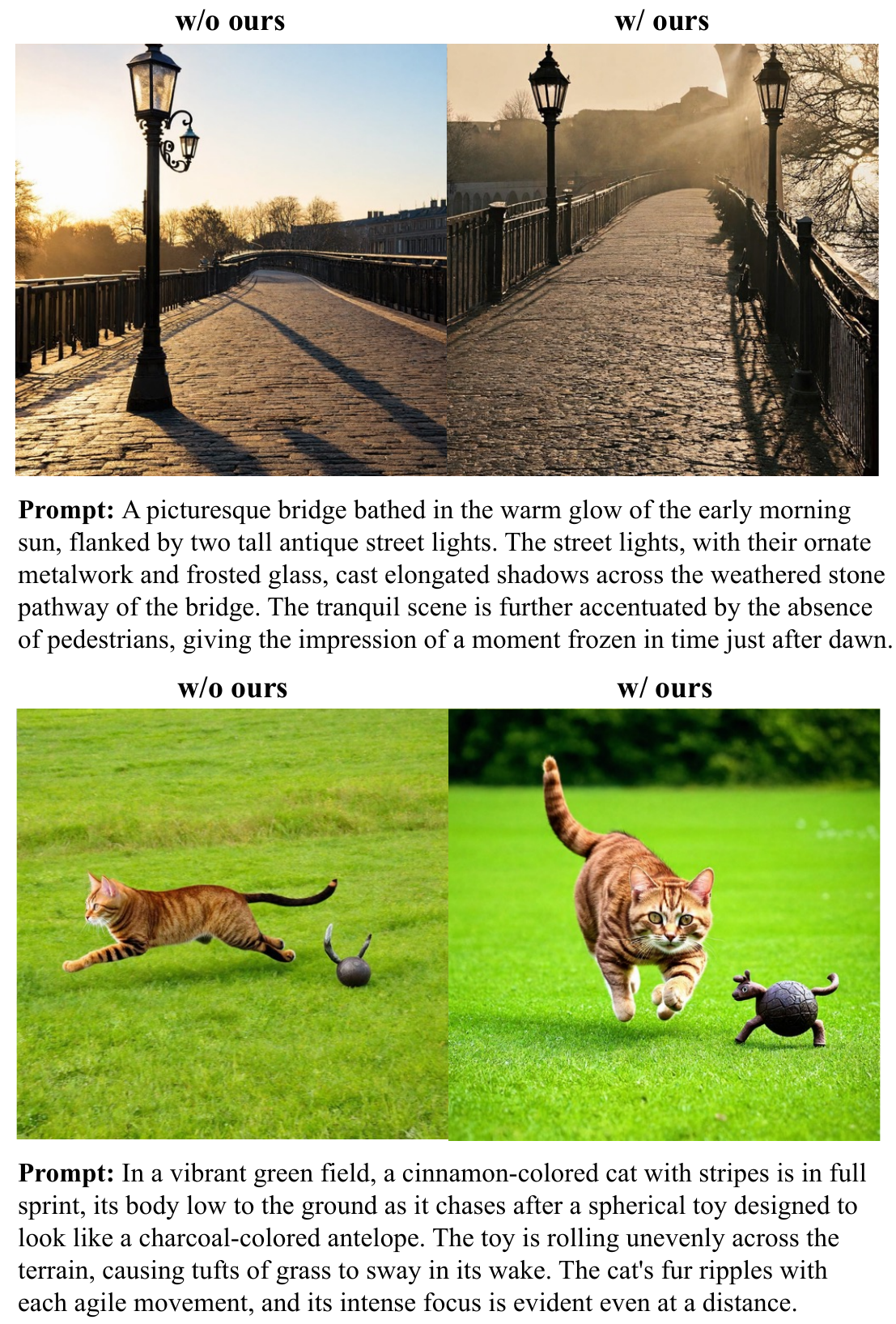}
    \caption{Qualitative comparison of image generation results on the DPG benchmark at 1024 × 1024 resolution. Compared with the baseline, ReMoA generates images that better align with the concepts described in the long text prompts.}
    \label{fig:dpg}
\end{figure}

\begin{table*}[t]
\centering
\small
\renewcommand{\arraystretch}{1.2}
\setlength{\tabcolsep}{5.5pt}
\vspace{5pt}
\caption{Evaluation results on the GenEval benchmark. ReMoA consistently outperforms both the SD3-medium baseline and its SFT variant, with the largest gains when targeting multiple layers. Best per column in \textbf{bold} and second-best \underline{underlined}.}
\begin{tabular}{l|cccccc|c}
\toprule
\rowcolor{CornflowerBlue!10}
\textbf{Model}& \textbf{Single object} & \textbf{Two object} & \textbf{Counting} & \textbf{Position} & \textbf{Colors} & \textbf{Color attr} & \textbf{Overall (\%)} \\
\hline
SD3-medium \cite{esser2024sd3} & 100.00 & 84.60 & 60.31 & 21.50 & 86.17 & 55.00 & 67.93  \\
SD3-medium + SFT  & 100.00 & 83.59 & 61.56 & 26.75 & 84.57 & 53.50 & 68.33 \\
ReMoA (Target on the middle layer) & 100.00 & 86.35 & 60.31 & 30.50 & 85.04 & 61.75 & \underline{70.66} \\
ReMoA (Target on multiple layers) & 100.00 & 89.65 & 64.38 & 21.75 & 84.31 & 61.00 & \textbf{71.18} \\
\hline
\multicolumn{6}{l}{Ablations} \\
\hline
ReMoA (w/o modulation) & 99.38 & 88.13 & 59.38 & 31.00 & 88.30 & 56.00 & 70.36 \\
ReMoA (w/o recursion)  & 99.38 & 84.85 & 59.06 & 34.00 & 86.97 & 55.00 & 69.88 \\
\bottomrule
\end{tabular}
\label{tab:geneval}
\vspace{15pt}

\centering
\small
\renewcommand{\arraystretch}{1.2}
\setlength{\tabcolsep}{8pt}
\caption{Evaluation results on the DPG benchmark. ReMoA achieves the best overall score, outperforming both SD3-medium and its SFT variant, with the largest gain in the global category. Best per column in \textbf{bold} and second-best \underline{underlined}.}
\begin{tabular}{l|ccccc|c}
\toprule
\rowcolor{CornflowerBlue!10}
\textbf{Model} & \textbf{Attribute} & \textbf{Entity} & \textbf{Relation} & \textbf{Global} & \textbf{Other} & \textbf{Overall (\%)} \\
\hline
SD3-medium \cite{esser2024sd3}  & 87.39 & 90.24& 92.15 & 85.71 & 80.40 & 87.18 \\ 
SD3-medium + SFT & 88.14 & 89.90  & 92.80 & 85.51 & 80.72 & \underline{87.41} \\ 
ReMoA (Target on multiple layers)& 88.34 & 90.39 & 91.53 & 88.15 & 82.80 & \textbf{88.25} \\ 
\bottomrule
\end{tabular}
\label{tab:dpg}
\end{table*}

\subsection{Text-to-image generation tasks}
\label{sec:t2i}

Computational depth induced by the recursion component is expected to be valuable, especially for integrating complex multimodal information such as text-to-image generation, which demands understanding compositional relationships and aligning them across vision and text. We further investigate the benefits of ReMoA in the GenEval and DPG benchmarks.

\paragraph{Evaluation on the GenEval benchmark}

To evaluate the recursion model's effectiveness in aligning vision tokens with more complex text prompts, we leverage the GenEval benchmark \cite{ghosh2023geneval} to assess the model's text-following ability. The GenEval benchmark \cite{ghosh2023geneval} measures text-following ability across compositional attributes including object count, color, position, and co-occurrence. Our method is implemented by replacing a single or multiple layers in the SD3-medium model and fine-tuning from pretrained SD3 weights. Table \ref{tab:geneval} shows that ReMoA consistently outperforms both the SD3-medium baseline and its SFT variant, improving the overall score from 68.33\% (SD3-medium + SFT) to 71.18\% when targeting multiple layers. Additionally, targeting only the middle layer reaches 70.66\%, suggesting that recursive refinement at a single well-chosen layer captures much of the benefit with fewer modified layers. 

To validate the design choices of the recursive and modular mechanisms, we conduct ablation studies on each component. Table \ref{tab:geneval} reports the corresponding results. Removing modulation (ReMoA w/o modulation) lowers the final score from 71.18\% to 70.36\%, and removing recursion (ReMoA w/o recursion) lowers it further to 69.88\%. These drops confirm that both the recursive and modular designs contribute meaningfully to the overall performance gains.

Furthermore, we present generated images at a resolution of 1024 × 1024 in Figure \ref{fig:sd3-compare}, showing that the recursive approach maintains high sample fidelity and improves text alignment over the SD3 baseline. In contrast, SD3 often fails to faithfully reflect details described in the prompts.

\paragraph{Discussion on the refinement method}

We study where to inject the frozen base model's output in the recursive refinement process. In particular, we compare the generation results when leveraging the base output at every step versus only at the final step (Figure \ref{fig:lora}). We find that using the frozen base output at the final step yields more stable training, whereas repeatedly adding it blurs the generated image details. A plausible explanation is that each pass compounds the smoothing effect, drifting the model toward over-smoothed predictions across the successive recursions.

\paragraph{Evaluation on the DPG benchmark}

We evaluate our method on the DPG benchmark using the multiple-layer variant with the same settings as in the GenEval evaluation. Table \ref{tab:dpg} presents the comparison results. Our method demonstrates enhanced semantic alignment, achieving an overall score of 88.25\% and outperforming both SD3-medium (87.18\%) and SD3-medium + SFT (87.41\%). The largest gain is on the global category (88.15\% vs. 80.40\%), while performance remains comparable on the remaining categories. We show generated samples from the DPG benchmark at a resolution of 1024 × 1024 in Figure \ref{fig:dpg}.

\section{Conclusions}

In this work, we proposed a recursive modular adaptation framework, ReMoA, for enhanced image generation in diffusion models while maintaining a lightweight architecture through a set of sparsely activated neural modules. Empirical experiments on both Diffusion Transformers and Stable Diffusion models demonstrate that inducing recursive modulation in the latent space benefits class-conditioned generation and text-following capability. Beyond quantitative gains, our analyses reveal that the neural modules learn to specialize across diffusion timesteps, with latent trajectories diverging at later, low-noise steps to support context-adaptive refinement. These findings suggest that recursive, sparsely routed computation offers a principled and efficient route to stronger cross-modal alignment in diffusion models, without inflating the parameter budget or altering the underlying backbone. We hope this work will contribute to the study of iterative latent refinement in multimodal models for better-aligned cross-modal generation.

\paragraph{Limitations}

Despite providing enhanced performance with lightweight adaptation, our method does not guarantee an optimal latent computation depth during longer recursions. For example, a halting mechanism that decides when to stop latent reasoning could enable more efficient computation. Moreover, although we demonstrate our method on vision diffusion models, its applicability to other modalities remains for future study.

\bibliographystyle{unsrt}
\bibliography{main}

@article{snell2024scaling,
  title={Scaling llm test-time compute optimally can be more effective than scaling model parameters},
  author={Snell, Charlie and Lee, Jaehoon and Xu, Kelvin and Kumar, Aviral},
  journal={arXiv preprint arXiv:2408.03314},
  year={2024}
}

@inproceedings{rombach2022high,
  title={High-resolution image synthesis with latent diffusion models},
  author={Rombach, Robin and Blattmann, Andreas and Lorenz, Dominik and Esser, Patrick and Ommer, Bj{\"o}rn},
  booktitle={Proceedings of the IEEE/CVF conference on computer vision and pattern recognition},
  pages={10684--10695},
  year={2022}
}

@article{jolicoeur2025less,
  title={Less is more: Recursive reasoning with tiny networks},
  author={Jolicoeur-Martineau, Alexia},
  journal={arXiv preprint arXiv:2510.04871},
  year={2025}
}

@article{zhao2024dynamic,
  title={Dynamic diffusion transformer},
  author={Zhao, Wangbo and Han, Yizeng and Tang, Jiasheng and Wang, Kai and Song, Yibing and Huang, Gao and Wang, Fan and You, Yang},
  journal={arXiv preprint arXiv:2410.03456},
  year={2024}
}

@article{dhariwal2021diffusion,
  title={Diffusion models beat gans on image synthesis},
  author={Dhariwal, Prafulla and Nichol, Alexander},
  journal={Advances in neural information processing systems},
  volume={34},
  pages={8780--8794},
  year={2021}
}

@article{teng2024dim,
  title={Dim: Diffusion mamba for efficient high-resolution image synthesis},
  author={Teng, Yao and Wu, Yue and Shi, Han and Ning, Xuefei and Dai, Guohao and Wang, Yu and Li, Zhenguo and Liu, Xihui},
  journal={arXiv preprint arXiv:2405.14224},
  year={2024}
}

@article{heusel2017gans,
  title={Gans trained by a two time-scale update rule converge to a local nash equilibrium},
  author={Heusel, Martin and Ramsauer, Hubert and Unterthiner, Thomas and Nessler, Bernhard and Hochreiter, Sepp},
  journal={Advances in neural information processing systems},
  volume={30},
  year={2017}
}

@article{salimans2016improved,
  title={Improved techniques for training gans},
  author={Salimans, Tim and Goodfellow, Ian and Zaremba, Wojciech and Cheung, Vicki and Radford, Alec and Chen, Xi},
  journal={Advances in neural information processing systems},
  volume={29},
  year={2016}
}

@article{nash2021generating,
  title={Generating images with sparse representations},
  author={Nash, Charlie and Menick, Jacob and Dieleman, Sander and Battaglia, Peter W},
  journal={arXiv preprint arXiv:2103.03841},
  year={2021}
}

@article{ghosh2023geneval,
  title={Geneval: An object-focused framework for evaluating text-to-image alignment},
  author={Ghosh, Dhruba and Hajishirzi, Hannaneh and Schmidt, Ludwig},
  journal={Advances in Neural Information Processing Systems},
  volume={36},
  pages={52132--52152},
  year={2023}
}

@article{zhao2025dydit++,
  title={DyDiT++: Dynamic Diffusion Transformers for Efficient Visual Generation},
  author={Zhao, Wangbo and Han, Yizeng and Tang, Jiasheng and Wang, Kai and Luo, Hao and Song, Yibing and Huang, Gao and Wang, Fan and You, Yang},
  journal={arXiv preprint arXiv:2504.06803},
  year={2025}
}

@article{pham2025multimodal,
  title={Multimodal chain of continuous thought for latent-space reasoning in vision-language models},
  author={Pham, Tan-Hanh and Ngo, Chris},
  journal={arXiv preprint arXiv:2508.12587},
  year={2025}
}

@article{ma2025cocova,
  title={CoCoVa: Chain of Continuous Vision-Language Thought for Latent Space Reasoning},
  author={Ma, Jizheng and Zhou, Xiaofei and Song, Yanlong and Yan, Han},
  journal={arXiv preprint arXiv:2511.02360},
  year={2025}
}

@article{zhu2025llada,
  title={LLaDA-MoE: A Sparse MoE Diffusion Language Model},
  author={Zhu, Fengqi and You, Zebin and Xing, Yipeng and Huang, Zenan and Liu, Lin and Zhuang, Yihong and Lu, Guoshan and Wang, Kangyu and Wang, Xudong and Wei, Lanning and others},
  journal={arXiv preprint arXiv:2509.24389},
  year={2025}
}

@article{madan2021fast,
  title={Fast and slow learning of recurrent independent mechanisms},
  author={Madan, Kanika and Ke, Nan Rosemary and Goyal, Anirudh and Sch{\"o}lkopf, Bernhard and Bengio, Yoshua},
  journal={arXiv preprint arXiv:2105.08710},
  year={2021}
}

@article{lei2024compete,
  title={Compete and compose: Learning independent mechanisms for modular world models},
  author={Lei, Anson and Nolte, Frederik and Sch{\"o}lkopf, Bernhard and Posner, Ingmar},
  journal={arXiv preprint arXiv:2404.15109},
  year={2024}
}

@article{wang2025hierarchical,
  title={Hierarchical Reasoning Model},
  author={Wang, Guan and Li, Jin and Sun, Yuhao and Chen, Xing and Liu, Changling and Wu, Yue and Lu, Meng and Song, Sen and Yadkori, Yasin Abbasi},
  journal={arXiv preprint arXiv:2506.21734},
  year={2025}
}

@article{wang2022rationale,
  title={Rationale-augmented ensembles in language models},
  author={Wang, Xuezhi and Wei, Jason and Schuurmans, Dale and Le, Quoc and Chi, Ed and Zhou, Denny},
  journal={arXiv preprint arXiv:2207.00747},
  year={2022}
}

@article{banino2021pondernet,
  title={Pondernet: Learning to ponder},
  author={Banino, Andrea and Balaguer, Jan and Blundell, Charles},
  journal={arXiv preprint arXiv:2107.05407},
  year={2021}
}

@misc{lin2015microsoftcococommonobjects,
      title={Microsoft COCO: Common Objects in Context}, 
      author={Tsung-Yi Lin and Michael Maire and Serge Belongie and Lubomir Bourdev and Ross Girshick and James Hays and Pietro Perona and Deva Ramanan and C. Lawrence Zitnick and Piotr Dollár},
      year={2015},
      eprint={1405.0312},
      archivePrefix={arXiv},
      primaryClass={cs.CV},
      url={https://arxiv.org/abs/1405.0312}, 
}

@inproceedings{sun2025associative,
  title={Associative transformer},
  author={Sun, Yuwei and Ochiai, Hideya and Wu, Zhirong and Lin, Stephen and Kanai, Ryota},
  booktitle={Proceedings of the Computer Vision and Pattern Recognition Conference},
  pages={4518--4527},
  year={2025}
}

@inproceedings{deng2009imagenet,
  title={Imagenet: A large-scale hierarchical image database},
  author={Deng, Jia and Dong, Wei and Socher, Richard and Li, Li-Jia and Li, Kai and Fei-Fei, Li},
  booktitle={2009 IEEE conference on computer vision and pattern recognition},
  pages={248--255},
  year={2009},
  organization={Ieee}
}

@inproceedings{bao2023all,
  title={All are worth words: A vit backbone for diffusion models},
  author={Bao, Fan and Nie, Shen and Xue, Kaiwen and Cao, Yue and Li, Chongxuan and Su, Hang and Zhu, Jun},
  booktitle={Proceedings of the IEEE/CVF conference on computer vision and pattern recognition},
  pages={22669--22679},
  year={2023}
}

@article{blum2021theoretical,
  title={A theoretical computer science perspective on consciousness},
  author={Blum, Manuel and Blum, Lenore},
  journal={Journal of Artificial Intelligence and Consciousness},
  volume={8},
  number={01},
  pages={1--42},
  year={2021},
  publisher={World Scientific}
}

@article{wei2022chain,
  title={Chain-of-thought prompting elicits reasoning in large language models},
  author={Wei, Jason and Wang, Xuezhi and Schuurmans, Dale and Bosma, Maarten and Xia, Fei and Chi, Ed and Le, Quoc V and Zhou, Denny and others},
  journal={Advances in neural information processing systems},
  volume={35},
  pages={24824--24837},
  year={2022}
}

@article{goyal2021coordination,
  title={Coordination among neural modules through a shared global workspace},
  author={Goyal, Anirudh and Didolkar, Aniket and Lamb, Alex and Badola, Kartikeya and Ke, Nan Rosemary and Rahaman, Nasim and Binas, Jonathan and Blundell, Charles and Mozer, Michael and Bengio, Yoshua},
  journal={arXiv preprint arXiv:2103.01197},
  year={2021}
}

@article{geiping2025scaling,
  title={Scaling up Test-Time Compute with Latent Reasoning: A Recurrent Depth Approach},
  author={Geiping, Jonas and McLeish, Sean and Jain, Neel and Kirchenbauer, John and Singh, Siddharth and Bartoldson, Brian R and Kailkhura, Bhavya and Bhatele, Abhinav and Goldstein, Tom},
  journal={arXiv preprint arXiv:2502.05171},
  year={2025}
}

@inproceedings{esser2024sd3,
  title={Scaling rectified flow transformers for high-resolution image synthesis},
  author={Esser, Patrick and Kulal, Sumith and Blattmann, Andreas and Entezari, Rahim and M{\"u}ller, Jonas and Saini, Harry and Levi, Yam and Lorenz, Dominik and Sauer, Axel and Boesel, Frederic and others},
  booktitle={International Conference on Machine Learning},
  year={2024}
}

@article{hu2024ella,
  title={Ella: Equip diffusion models with llm for enhanced semantic alignment},
  author={Hu, Xiwei and Wang, Rui and Fang, Yixiao and Fu, Bin and Cheng, Pei and Yu, Gang},
  journal={arXiv preprint arXiv:2403.05135},
  year={2024}
}

@inproceedings{peebles2023DIT,
  title={Scalable diffusion models with transformers},
  author={Peebles, William and Xie, Saining},
  booktitle={Proceedings of the IEEE/CVF International Conference on Computer Vision},
  pages={4195--4205},
  year={2023}
}

@article{chen2023pixartalpha,
  title={Pixart-$alpha$: Fast training of diffusion transformer for photorealistic text-to-image synthesis},
  author={Chen, Junsong and Yu, Jincheng and Ge, Chongjian and Yao, Lewei and Xie, Enze and Wu, Yue and Wang, Zhongdao and Kwok, James and Luo, Ping and Lu, Huchuan and others},
  journal={arXiv preprint arXiv:2310.00426},
  year={2023}
}

@inproceedings{chen2024pixartsigma,
  title={Pixart-$\sigma$: Weak-to-strong training of diffusion transformer for 4k text-to-image generation},
  author={Chen, Junsong and Ge, Chongjian and Xie, Enze and Wu, Yue and Yao, Lewei and Ren, Xiaozhe and Wang, Zhongdao and Luo, Ping and Lu, Huchuan and Li, Zhenguo},
  booktitle={European Conference on Computer Vision},
  pages={74--91},
  year={2024},
  organization={Springer}
}

@article{ho2020denoising,
  title={Denoising diffusion probabilistic models},
  author={Ho, Jonathan and Jain, Ajay and Abbeel, Pieter},
  journal={Advances in Neural Information Processing Systems},
  volume={33},
  pages={6840--6851},
  year={2020}
}

@article{song2020denoising,
  title={Denoising diffusion implicit models},
  author={Song, Jiaming and Meng, Chenlin and Ermon, Stefano},
  journal={arXiv preprint arXiv:2010.02502},
  year={2020}
}

@inproceedings{zhou2025goldennoisediffusionmodels,
  title={Golden noise for diffusion models: A learning framework},
  author={Zhou, Zikai and Shao, Shitong and Bai, Lichen and Zhang, Shufei and Xu, Zhiqiang and Han, Bo and Xie, Zeke},
  booktitle={Proceedings of the IEEE/CVF International Conference on Computer Vision},
  pages={17688--17697},
  year={2025}
}

@article{hu2021lora,
  title={Lora: Low-rank adaptation of large language models},
  author={Hu, Edward J and Shen, Yelong and Wallis, Phillip and Allen-Zhu, Zeyuan and Li, Yuanzhi and Wang, Shean and Wang, Lu and Chen, Weizhu},
  journal={arXiv preprint arXiv:2106.09685},
  year={2021}
}

\end{document}